\documentclass[11pt,a4paper]{article}
\usepackage[hyperref]{eacl2021}
\usepackage{times}
\usepackage{latexsym}
\usepackage{booktabs}
\usepackage{multicol}
\usepackage{multirow}
\usepackage{pbox}
\usepackage{amsmath}

\usepackage{microtype}

\aclfinalcopy 
\def\aclpaperid{19} 

\title{A Computational Analysis of Vagueness in Revisions of Instructional Texts}

\author{Alok Debnath \\
  Kohli Center for Intelligent Systems\\
  International Institute of Information\\
  Technology, Hyderabad\\
  \texttt{alok.debnath@research.iiit.ac.in} \\\And
  Michael Roth \\
  Institute for Natural Language Processing \\
  University of Stuttgart \\
  \texttt{michael.roth@ims.uni-stuttgart.de} \\
}

\date{}

\begin{document}
\maketitle
\begin{abstract}
     \emph{WikiHow} is an open-domain repository of instructional articles for a variety of tasks, which can be revised by users. 
     In this paper, we extract pairwise versions of an instruction before and after a revision was made. 
     Starting from a noisy dataset of revision histories, we specifically extract and analyze edits that involve cases of vagueness in instructions.
     We further investigate the ability of a neural model to distinguish between two versions of an instruction in our data by adopting a pairwise ranking task from previous work and showing improvements over existing baselines.
\end{abstract}

\section{Introduction}

Instructional texts aim to describe the actions necessary to accomplish a task or goal, in as clear and concise a manner as possible. 
\textit{WikiHow}\footnote{\url{https://www.wikihow.com/}} is an extensive compendium of instructional guides for various
topics and domains. Any user may edit the articles, and \textit{WikiHow} collates these revision histories. The edit history of such informal instructional articles is a source of user-generated data that can help identify possible reasons and necessities for editing. \emph{wikiHowToImprove} \citep{anthonio2020wikihowtoimprove} is a dataset that compiles revision histories for the analysis of linguistic phenomena that occur in edits of instructional texts, ranging from the correction of typos and grammatical errors to the clarification of ambiguity and vagueness.

\begin{table}[t]
    \centering
    \begin{tabular}{@{}p{3.60cm}p{3.60cm}@{}}
    \toprule
        \bf Original Sentence & \bf Revised Sentence \\
    \midrule
        \pbox{3.60cm}{Then, \textbf{make} the floor and walls of your house.} &  \pbox{3.60cm}{Then, \textbf{design} the floor and walls of your house.} \\ \midrule
        \pbox{3.58cm}{When you \textbf{go} to the Hogwarts park...} & \pbox{3.58cm}{When you \textbf{visit} the Hogwarts park...} \\ \midrule
        \pbox{3.58cm}{\textbf{Get} a flexible single cord.} & \pbox{3.58cm}{\textbf{Purchase} a flexible single cord.} \\ \toprule
    \end{tabular}
    \caption{Examples of vague instructions and their more clarified versions from the \emph{wikiHowToImprove} Dataset}
    \label{tab:example}
\end{table}

In this paper, we focus on cases of lexical vagueness, defined as ``lexeme[s] with a single but nonspecific meaning'' \cite{tuggy1993ambiguity}, which can potentially cause misunderstandings in instructional texts. Specifically, we study vagueness based on the change in the main verb in the original and revised version of an instruction. We say that an instruction was vague if, upon revision, the revised main verb is contextually more specific than the original version.
Some examples of vague and clarified instructions are provided in Table \ref{tab:example}. As indicated by the examples, the revised verb is usually more specific in that it provides additional information on how or why an action needs to be taken.

The classification of vague and clarified instructions is a first step towards automatic text editing for clarification based on linguistic criteria such as ambiguity and vagueness at a sentence level. Existing tools for text editing focus on text simplification and fact editing \citep{malmi2019encode}, while others are designed for grammatical error correction \citep{xie2018noising}. Our work acts as the first step towards automated editing based on linguistic criteria by identifying vague instructions and differentiating them from ``clarified" ones. Our use of the \emph{wikiHowToImprove} corpus also utilizes a resource of edit pairs, therefore introducing a new dataset for the linguistic study of vagueness as well as exploring the general versatility of such corpora.

Our contributions are to create a dataset of vague and clarified instructions, provide an analysis based on semantic frames, and demonstrate the first results of a neural model's ability to distinguish the two versions. We create and analyze the dataset by extracting relevant instances from \emph{wikiHowToImprove}, using POS tags, dependency features, and edit distance as constraints, as well as FrameNet frames as features (Section \ref{sec: data}). 
We then devise a pairwise ranking task, where we train and evaluate different neural models and analyze their performance based on frame relations and differences in distributional word representations (Section \ref{sec: experiments}).

\section{Related Work}

Our paper focuses on revisions in wikiHow for a specific linguistic phenomenon, namely vagueness. The motivation to use revision histories as corpora for NLP tasks was introduced by \citet{ferschke2013survey}. The task of defining and categorizing edit intentions has been explored well for the Wikipedia edits corpus \citep{yang2016edit, yang2017identifying}. More recently, \citet{anthonio2020wikihowtoimprove} performed a similar categorization on the revisions in \textit{WikiHow}.

Traditional computational analyses of vague statements have been based on logical representations \citep{devault2004interpreting, tang2008collective}. In contrast, our focus is on vagueness in terms of lexical changes in revisions, which is more similar to previous analyses that considered the context-dependent resolution of vague expressions such as colour references \citep{meo2014generating}. Other computational approaches to vagueness include, the detection of vague sense definitions in ontological resources \citep{alexopoulos2014vague} and website privacy policies \citep{lebanoff2018automatic} as well as the verification of historical documents \citep{vertan2019modelling}.

Our approach to identifying and classifying vagueness is analyzed using FrameNet frames which provide specialized relations among conceptual categories, in a manner similar to recent advances in neural models that use sentence-level information to perform hyponymy--hypernymy classification. \citet{roller2018hearst} analyzes lexico-syntactic pattern-based instances of word-specific hypernymy-hyponymy constructions. \citet{snow2004learning} explores the extraction of predefined patterns for hypernyms and hyponyms in the same sentence, while \citet{shwartz2016improving} incorporates distributional methods for their classification using sentence-level features. 

\section{Data Creation, Preprocessing, and Analysis}
\label{sec: data}

\emph{WikiHow} articles mostly contain instructions, but also include descriptions, explanations, and other non-instructional sentences that provide additional context. The \emph{wikiHowToImprove} corpus \cite{anthonio2020wikihowtoimprove} is an unfiltered corpus of revision histories. Therefore, we first need to extract those revisions where the original and revised versions are both instructional sentences, which can be done based on syntactic properties (\S \ref{ssec: cleaning}). We then use a FrameNet parser to determine the frames (and their relationships) evoked by the root verb in the original and revised version of an instruction (\S \ref{ssec: framenet}).

The final extracted data consists of only those revisions where the root verb has been modified to be more specific to the sentence. This extracted corpus consists of 41,615 sentences.

\subsection{Data Extraction and Cleaning}
\label{ssec: cleaning}

\emph{wikiHowToImprove} is a noisy source of data with misspellings, non-standard abbreviations, grammatical errors, emoticons, etc. In order to use the data for our task, we first perform some cleaning and preprocessing. 

We filter the typos and misspellings in the dataset by comparing all the vocabulary words to words in the English dictionary using the \textit{Enchant} python API\footnote{\url{https://pyenchant.github.io/}}. After filtering the typos, we POS tag and dependency parse the data using the \textit{Stanza} library\footnote{\url{https://stanfordnlp.github.io/stanza/}} \citep{qi2020stanza}. We discard all sentence pairs where the sentences are shorter than four or longer than 50 words.

We then create a sub-corpus of instructional sentneces by extracting those edit pairs in which both the original and revised version of a sentence fulfill at least one of the following criteria:
\begin{itemize}
    \item imperative form---the root verb has no nominal subject (e.g. ``Please finish the task'');
    \item instructional indicative form---the nominal subject of the root verb is `you,' `it' or `one' (e.g.
    ``You should finish the task'');
    \item passive form with `let'---the sentence is in passive voice, and the root verb is `let' (e.g. ``Let the paper be put on the table.'').
\end{itemize}

Finally, we retain only those sentence pairs whose character edit distance is smaller than 10. This filter was added after empirical tests to accommodate changes in the verb form and syntactic frame while ensuring that there are little to no additional edits (often just vandalism or spam).

\subsection{Verb Frame Analysis}
\label{ssec: framenet}

We perform an analysis of verb frame relations from this extracted corpus using the FrameNet hierarchy \citep{baker1998berkeley}. In order to identify evoked frames from the data, we use the INCEpTION Project's neural \emph{FrameNet Tools} parser\footnote{\url{https://github.com/inception-project/framenet-tools}} \citep{tubiblio106270, andre_markard_2020_3993970}. FrameNet Tools identifies the frame-evoking elements, the evoked frames, and the context elements' roles in these frame for a given sentence. 
In this work, we ignore role assignments and only consider predictions of evoked frames, which we found to be generally reliable in our data.\footnote{Although automatic frame identification is noisy, the tools used here are implementations of the unimodal model presented in \citet{botschen-etal-2018-multimodal}, which achieves a high accuracy of over 88\%.}

We extract the frame of the root verb in the original and revised sentences. For each pair, we identify the frame relation, if any, using the NLTK FrameNet API\footnote{\url{http://www.nltk.org/howto/framenet.html}}\citep{schneider2017nltk}. 
We found that most edits could be categorized into one of the following frame relations between the frames evoked by the original and revised verb frames:

\begin{enumerate}
    \item \textbf{Subframe-of}: The original frame refers to a complex scenario that consists of several individual states, one of which is the revised frame. (e.g. \textsc{Traversing}$\rightarrow$\textsc{Arriving}: ``\emph{Go} to the thumbs up log." is revised to ``\emph{Visit} the thumbs up log.") 
    
    \item \textbf{Inherits-from}: The frame of the revised verb elaborates on the frame evoked by the original verb (e.g., \textsc{Deciding$\rightarrow$Choosing}:
    ``\emph{Determine} the card you want to buy" is revised to ``\emph{Choose} which card you want to buy.")
    
    \item \textbf{Uses}: The frame of the revised verb uses or weakly inherits properties of the original verb frame (e.g., \textsc{Perception\_active$\rightarrow$Scrutiny}: ``\emph{Look} for the best fit for your taste" is revised to ``\emph{Search} for the best fit for your taste.").
\end{enumerate}

We also find cases of contextually relevant clarifications for phrasal verbs, such as ``\emph{Make} your bed" vs.\ ``\emph{Fix} your bed\ldots" which are not covered in FrameNet. Further, there are cases in which the FrameNet Tools parser did not identify the main verb or could not assign a frame.
For instance, the verb \emph{compel} as in ``you may feel \emph{compelled} \ldots" is not in FrameNet. We categorize these instances, which are fewer in number than the other categories, under a single \textbf{Other} category and leave further inspection to future work.
A distribution of instances over categories is shown in Table \ref{tab: stats}. Apart from instances from the `Other' category, we indeed found the main verbs in the revised versions of a sentence to be more specific than in the original versions.

\begin{table}[t]
    \centering
    \begin{tabular}{lrrrr}
        \toprule
        Relation        & \bf Total & \bf Train     & \bf Test  & \bf Val \\ 
        \midrule
        Usage           & 15,243    & 11,084        & 2,194     & 1,965  \\
        Inheritance     & 13,166    &  9,179        & 2,008     & 1,793  \\
        Subframe        &  9,481    &  6,835        & 1,720     &   926  \\
        Other           &  3,925    &  2,833        &   649     &   443  \\
        \midrule
        Total  & 41,615    & 30,044        & 6,237     & 5,334 \\
        \bottomrule
    \end{tabular}
    \caption{Number of sentences in the extracted dataset and distribution of FrameNet relations between original and revised verbs. We also show the distribution of train, test and validation for each frame relation.}
    \label{tab: stats}
\end{table}

\section{Pairwise Ranking Experiments}
\label{sec: experiments}

In this section, we investigate if a neural model can distinguish between the original and revised version of the same instruction. We describe a neural architecture that uses a joint representation designed for comparing two versions of a sentence before predicting an output. We compare our results to a standard BiLSTM-Attention model used in previous work \citep{anthonio2020wikihowtoimprove}. 

\subsection{System and Training Details}

The initial components of our system are two BiLSTM modules, LSTM$_{1A}$ and LSTM$_{1B}$, that each takes one version of a sentence as input. The individual BiLSTMs are followed by a joint layer LSTM$_{AB}$ and an additional layer of BiLSTM modules, LSTM$_{2A}$ and LSTM$_{2B}$, that re-encode the sentence based on the joint representations. The final layer is trained to predict for each sentence, whether it is the original or revised version, labeling them $0$ or $1$, respectively. 

In practice, we first encode versions $A$ and $B$ of an instruction using FastText embeddings or BERT. The embedded sentences $S_A$ and $S_B$ are then passed through LSTM$_{1A}$ and LSTM$_{1B}$ one (sub-word) token at a time. The hidden layers $\textbf{h}_{1A}$ and $\textbf{h}_{1B}$ are then concatenated and passed through LSTM$_{AB}$, whose output $\textbf{h}_{AB}$ is then concatenated again with the original hidden states to re-encode each sentence version in LSTM$_{2A}$ and LSTM$_{2B}$. Lastly a classification layer, trained using a cross-entropy objective, transforms the final representations $\textbf{h}_{2A}$ and $\textbf{h}_{2B}$ into a real-valued output score using self-attention, which is normalized by softmax and rounded to $\{0,1\}$. The equations below give a simplified summary of our implementation.\footnote{We will make the code available upon publication.}
\begin{align}
    & \textbf{h}_{1A} & = &\text{LSTM}_{1A} (S_A) \\
    & \textbf{h}_{1B} & = &\text{LSTM}_{1B} (S_B) \\
    & \textbf{h}_{AB} & = &\text{LSTM}_{AB} (\textbf{h}_{1A} \cdot \textbf{h}_{1B}) \\
    & \textbf{h}_{2A} & = &\text{LSTM}_{2A} (\textbf{h}_{AB} \cdot \textbf{h}_{1A}) \\
    & \textbf{h}_{2B} & = &\text{LSTM}_{2B} (\textbf{h}_{AB} \cdot \textbf{h}_{1B}) \\
    & l_A    & = & 
    \left[\frac{\exp(\textbf{w}^\top \textbf{h}_{2A})}{\exp(\textbf{w}^\top \textbf{h}_{2A}) + \exp(\textbf{w}^\top \textbf{h}_{2B})}\right]\\
    & l_B    & = & 
    \left[\frac{\exp(\textbf{w}^\top \textbf{h}_{2B})}{\exp(\textbf{w}^\top \textbf{h}_{2A}) + \exp(\textbf{w}^\top \textbf{h}_{2B})}\right]
\end{align}

\paragraph{Training Details} We experiment with both FastText \citep{grave2018learning} and BERT \citep{devlin2019bert}, using representations with a dimensionality of 300 components. The BiLSTMs modules $\text{LSTM}_{1A}, \text{LSTM}_{1B}, \text{LSTM}_{2A}$ and $\text{LSTM}_{2B}$ each comprise one hidden layer with 256 components, whereas the joint $\text{LSTM}_{AB}$ comprises one layer with 512 components. We train for 5 epochs with a batch size of 32 and a learning rate of $10^{-5}$.
The model is trained with a dropout of 0.2 for regularization. No dropout is applied to any BiLSTM layers or the self-attention layer.

For training, development, and testing, we split our data according to the existing partition given in \emph{wikiHowToImprove}.\footnote{\url{https://github.com/irshadbhat/wikiHowToImprove}} The resulting split consists of 30,044 sentence revision pairs in the training set, 6,237 pairs in the test set, and 5,334 pairs in the validation set.

\subsection{Results and Discussion}

\begin{table}[t]
    \centering
    \begin{tabular}{ccc}
    \toprule
        \bf Model Description                       & \bf Dataset   & \bf Accuracy \\ \midrule
        \citet{anthonio2020wikihowtoimprove}        & Entire        & 74.50\% \\ \midrule
        \citet{anthonio2020wikihowtoimprove}        & Filtered      & 64.08\% \\
        Our Model + FastText                        & Filtered      & 71.16\% \\
        Our Model + BERT                            & Filtered      & \bf 78.40\% \\ \toprule
    \end{tabular}
    \caption{Results of the pairwise ranking task, on the full wikiHowToImprove dataset (Entire) and our subset of instructional sentences (Filtered). 
    }
    \label{tab: classification}
\end{table}

Table \ref{tab: classification} shows the results of the pairwise ranking task. We find that our proposed model with BERT embeddings is the most accurate model for this task by a margin of about 7\%. We compare our results against the baseline provided by \citet{anthonio2020wikihowtoimprove}, which also makes use of ranking and a BiLSTM architecture. In contrast to our model, their baseline is a simple BiLSTM-Attention classification model using FastText embeddings. It does not use an intermediate joint representation to compare representations of two versions of an instruction. The baseline model has the advantage of being trained on individual sentences, but the increase in model accuracy for training sentence pairs by sharing context highlights the efficacy of the training regime. 

Their model provides an accuracy of about 64.08\% when trained and evaluated on the filtered corpus. Our model with FastText embeddings achieves an accuracy of 71.16\% ($+7.08\%$), which shows the relative importance of the joint representation.

\begin{table}[t]
    \centering
    \begin{tabular}{cl@{}}
    \toprule
        \bf Frame Relation & \bf Sentence Pair \\ 
        (\#errors / total) & \\ \toprule
        
        \multirow{1}{*}{Usage}          & \textbf{Make} a comic in Flash \\
        (503 / 1,965)                               & \textbf{Create} a comic in Flash\\ \midrule
        \multirow{1}{*}{Inheritance }    & \textbf{Check} the ``made in" label \\
        (352 / 1,793)                                & \textbf{Inspect} the ``made in" label \\ \midrule
        \multirow{1}{*}{Subframe}       & \textbf{Let} your hair dry\\
        (137 / 926)                                & \textbf{Allow} your hair to dry\\ \midrule
        \multirow{1}{*}{Other}          & Next, \textbf{try} to sneak out...\\
        (160 / 443)                                & Next, \textbf{attempt} to sneak out...\\ \toprule
    \end{tabular}
    \caption{Some examples of sentences which our BERT-based classifier could not distinguish between the original (top) and revised (bottom) versions. We find that confusable verbs (marked in bold) are mostly synonymous. The error and total counts from the validation set are provided in parenthesis for each relation type.}
    \label{tab: fallacies}
\end{table}

\paragraph{Discussion} We find that version pairs that involve a subframe relation are the easiest to distinguish across our model using both FastText and BERT, while pairs involving the usage relation are most often confused. The model using BERT embeddings performs better than the FastText-based model on revisions that do not involve any frame-to-frame relations according to FrameNet (referred to as `other' in Table~\ref{tab: stats}). 

In Table~\ref{tab: fallacies}, we provide examples where the model failed using both FastText and BERT. We observe that the models fail to correctly distinguish between sentences when the main verbs are synonymous. The embeddings of the most commonly confused verb pairs, which include $\langle$allow, permit$\rangle$, $\langle$choose, decide$\rangle$ and $\langle$create, make$\rangle$, have a cosine similarity of $0.8$ or higher, while the average cosine similarity between the representation of verb pairs is $0.47$. This insight shows that embeddings by themselves might be insufficient for this classification task. In future work, we will explore additional features such as indicator features derived from the discourse context (e.g., the position of a sentence) and from the FrameNet resource (e.g., properties of the frames evoked in a sentence).

\section{Conclusion}

In this paper, we extracted a corpus of clarifications of instructions from the \emph{wikiHowToImprove} corpus. We described a methodology for extracting version pairs of a sentence that are both instructional. We then identified cases in which a revision has clarified a vague instruction by analyzing the relationship between the frames evoked by the `original' verb and the `revised' verb.

In our experiments, we adopted a simple pairwise ranking task, in the same vein as performed by \citet{anthonio2020wikihowtoimprove} on the entire \emph{wikiHowToImprove} dataset. We extended a simple BiLSTM architecture with a joint component and explored different embeddings methods, observing that both modifications lead to improvements over baselines presented in previous work.

We hope that our methodology of extracting linguistically interesting cases of revisions from a noisy dataset can be extended to more phenomena and other corpora in future work. This direction has the potential of paving the way for developing automated revision and editing methods beyond typo, style, and grammar correction.

\ifaclfinal

\section*{Acknowledgements}

The research presented in this paper was funded by the DFG Emmy Noether program (RO 4848/2-1).

\fi

\bibliography{eacl2021}
\bibliographystyle{acl_natbib}


\end{document}